# Does the testing environment matter? Carsickness across on-road, test-track, and driving simulator conditions

Georgios Papaioannou[1], Barys Shyrokau[1]

[1]Cognitive Robotics, Delft University of Technology, 2628 CD, Delft, The Netherlands

**ABSTRACT**

Carsickness has gained significant attention with the rise of automated vehicles, prompting extensive research across on-road, test-track, and driving simulator environments to understand its occurrence and develop mitigation strategies. However, the lack of carsickness standardization complicates comparisons across studies and environments. Previous works demonstrate measurement validity between two setups at most (e.g., on-road vs. driving simulator), leaving gaps in multi-environment comparisons. This study investigates the recreation of an on-road motion sickness exposure - previously replicated on a test track - using a motion-based driving simulator. Twenty-eight participants performed an eyes-off-road non-driving task while reporting motion sickness using the Misery Scale during the experiment and the Motion Sickness Assessment Questionnaire afterward. Psychological factors known to influence motion sickness were also assessed. The results present subjective and objective measurements for motion sickness across the considered environments. In this paper, acceleration measurements, objective metrics and subjective motion sickness ratings across environments are compared, highlighting key differences in sickness occurrence for simulator-based research validity. Significantly lower motion sickness scores are reported in the simulator compared to on-road and test-track conditions, due to its limited working envelope to reproduce low-frequency (<0.5 Hz) motions, which are the most provocative for motion sickness.

## INTRODUCTION

The engagement in non-driving related tasks (NDRT) is expected to be at the forefront for the wide acceptance of automated vehicles. However, all the envisaged designs of AVs are expected to provoke carsickness and discomfort. Yet, major challenges remain to address carsickness in AVs. Extensive research with human experiments is being carried out in different testing environments to understand the occurrence of carsickness and eventually develop countermeasures to mitigate it in the context of AVs (Papaioannou et al. (2025)). However, comparison studies across testing environment are scarce, and limit the standardization of carsickness studies (Bos et al. (2022)), making eventually difficult to compare results between different testing environments. This study focuses on comparing carsickness across testing environments.

Testing environments for carsickness need to balance two requirements: replicability and realism. The motion stimulus shall be maintained as similar as possible across participants and conditions, especially different countermeasures are tested or passenger monitoring/observation over the ride is conducted.

Common testing environments for carsickness research are driving simulators, test-tracks, or public roads, which achieve different compromises among these requirements. Simulators even employed with advanced motion cueing algorithms (Khusro et al. (2020)), have been shown to provoke lower MS levels compared to on-road driving (Dam et al. (2024), Mühlbacher et al. (2020), Talsma et al. (2023)). However, they provide high replicability and repeatability within-subjects. On-road and test-track experiments assess carsickness more realistically, but other disadvantages occur. Even with trained drivers or fully automated vehicles, driving behavior in public road studies is not consistently reproducible due to unexpected dynamic events (vulnerable road users, other vehicles, traffic lights etc.). Despite their high-risk, on-road studies are very costly and risky due to accidents probability. At the same time, test-tracks are a feasible solution since they allow higher replicability than on-road studies. However, test-tracks are not accessible to all researchers while they are relatively expensive. Recently, building upon motion planning methods to mitigate motion sickness (Jain et al., 2023, Htike et al. 2021), Harmankaya et al (2025) developed a method which effectively replicated on-road sickness exposure in a compact test track. This method aimed to cancel the costs and the need for large test-tracks for motion sickness experiments. Pham Xuan et al. (2025) used this method to compare the methodological aspects of carsickness assessment and its modulating factors in two different testing environments (test-track and on-road). This paper extends this work further by exploring the carsickness assessment in a driving simulator compared to on-road and test-track.

One of the primary advantages of simulators is their capability to reproduce scenarios that are difficult or unsafe to evaluate in real-world driving conditions. They are intensively used for investigating vehicle dynamics, human-machine interaction, and driver behavior under controlled and repeatable conditions. For example, for steering feel evaluation, allowing detailed assessment of steering system tuning, on-center behavior, and subjective handling characteristics without the need for extensive road testing (Shyrokau, 2018). They also play a key role in the development and validation of advanced driver assistance systems, enabling safe testing of critical scenarios and control strategies (Rossi, 2020). Over the last decade, driving simulators have increasingly been adopted for motion sickness and comfort studies (Aykent, 2014), as they provide a controlled environment for systematic investigation of sensory conflicts and human perception. within a safe, controlled, and repeatable environment.

This paper is structured as follows: firstly, the methods of the experiment are explained, secondly the results are presented and discussed, and finally conclusions are extracted.

**METHODS**

Ethics statement: The study is approved by the Human Research Ethics Council of Delft University of Technology (Delft, The Netherlands; application number 5751). All participants gave their written informed consent before participation in the study.

Stimulus: The experiment was conducted in Delft Advanced Vehicle Simulator (DAVSi) (Khusro et al. (2020)) (Figure 1). The on-road drive recorded by Harmankaya et al. (2025) was replicated in DAVSi, by providing the longitudinal and lateral accelerations from the on-road experiment. DAVSi translated these to feasible accelerations within the simulator using an adaptive washout filter-based Motion Cueing Algorithm (MCA) [E2M Technologies, 2019]. The accelerations from the on-road, the test-track (as resulted by Harmankaya's et al. (2025) method), and the simulator are presented and compared in Figure 2, as analysed in the frequency domain. The acceleration spectrum in the simulator signal exhibits lower amplitude in the lower frequencies, which are the most provocative for motion sickness.

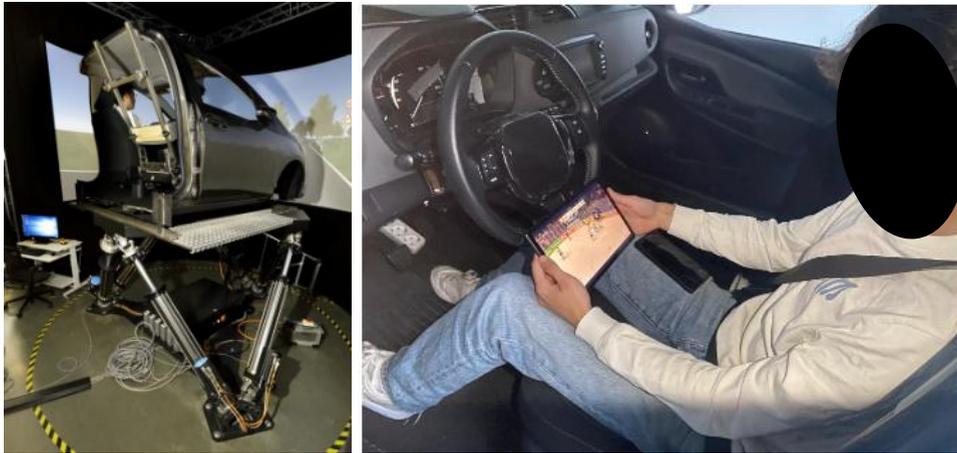

**Figure 1:** (left) Delft Advanced Vehicle Simulator (DAVSi) (right) Participants' posture while being driven in the experiment.

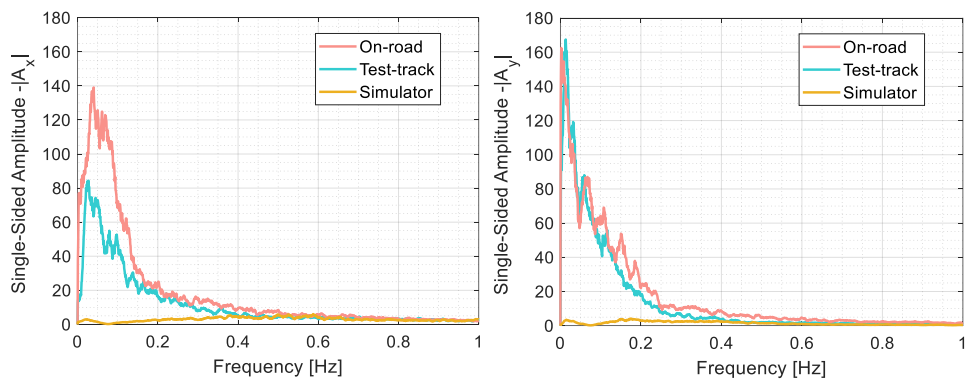

**Figure 2:** Power spectral density [m/s2Hz] of unweighted (left) longitudinal (Ax), and (right) lateral accelerations (Ay).

Procedure: Prior to the experiment, the participants were informed about the general aim of the study and gave their written consent. Then, participants filled out a pre-drive questionnaire consisting of: Part 1: Anthropometrics, Part 2: Self-Assessment Manikin (SAM) questionnaire (Bradley and Lang, (1994)), and Part 3: Motion sickness assessment questionnaire (MSAQ) (Gianaros et al., 2001). Thereafter, the experimental session commenced. Participants gave their carsickness level on the MISC every minute. After reaching the end of the path or after stating a MISC level of 6 or higher, the experiment was terminated.

Thereafter, participants filled out a post-drive questionnaire 3, which consisted of Part 1: SAM questionnaire, Part 2: Comfort (2.1 MSAQ, 2.2 Acceptance (Van Der Laan et al., 1997), 2.3 ARCA (Marberger et al., 2022)), Part 3: Trust and Part 4: Perceived Safety (Nordhoff et al., 2021). More details about the questionnaires are provided in detailed in Pham Xuan et al. (2025), whose experimental design was replicated in this work with minor modifications.

All participants were engaged in a video watching activity. More specifically, we used two videos from the same genre on a tablet held with the hands on the lap. Both videos depicted sporting events (a tennis and an ice-hockey game) to arouse similar emotional levels. To ensure and assess the participants engagement on the NDRTs, the videos were edited and included unexpected events. More specifically, a ball overlapped the tennis ball or hockey puck at random moments. The participants were required to count these events to verify their engagement in the NDRT. In total, sixteen events took place in both videos until the 25th minute.

Participants: For the selection of participants, the short Motion Sickness Susceptibility Questionnaire (MSSQ) is used. The overall MSSQ-Short Score as well as the item regarding the experience of carsickness in the past ten years were weighted equally to assess the current theoretical susceptibility to carsickness. Depending on this score, participants were assigned to one of five categories, as described in Pham Xuan (2023).

Participants from categories B, C, D and E are invited to the experiment. Gender balanced is attempted. The average susceptibility of simulator participants is 12.4 ± 10.4, compared to 13.0 ± 6.0 for test-track/on-road participants. The mean MSSQ scores between the experiments showed no significance using the Welch t-test ($p$=0.783). In the current study, 28 participants joined, of which 15 identify as male, 11 female and 2 non-binary/third gender. The mean age of the participants was 22 ± 2.2 years. The on-road and test-track experiments had the same group of participants, consisting of 47 participants, of whom 29 identify as male, 17 female and 1 non-binary/third gender. The mean age of the participants in this study was 28 ± 13 years. Their focus level over the NDRT during the ride was similar in both conditions and relatively high. More specifically, the participants counted on average 16.54 out of 20 events in total during experiment.

Dependent variables: To measure carsickness during the actual experimental rides, the misery scale (MISC by Bos et al., 2005) was used. Participants were asked to verbally give their subjective motion sickness rating on the MISC scale once every minute. If participants reached a value equal to 6 (little nausea) or higher, the experimental session was terminated immediately. This was done due to ethical concerns. To analyse the MISC ratings between the conditions, the same procedure as Pham Xuan et al. (2025) was implemented.

Statistical analysis: A mixed-design ANOVA was conducted to assess the effects of the experimental condition (Between-Subjects factor: Simulator vs. Test-track, Simulator vs. On-road) and time (Within-Subjects factor: 0–25 minutes) on MISC scores. Mauchly's test was used to evaluate the assumption of sphericity in the MISC Data. As the assumption was violated ($p < 0.05$), Greenhouse-Geisser corrections were applied to the degrees of freedom for the within-subjects' effects.

To identify the specific onset of divergence between conditions, post-hoc analysis was performed using series of independent Mann-Whitney U tests (rank-sum tests) at each minute. This non-parametric approach was selected due to the ordinal nature and non-normal distribution of MISC scores. Statistical significance was defined as $p < 0.05$, with multi-level significance indicated as follows: *$p < 0.05$, **$p < 0.01$, and ***$p < 0.001$.

## Results

**Objective data analysis**: For, the comparison of the two experiments, we explore the calculation of Motion Sickness Dose Value (MSDV) based on ISO-2631 (1997) and similarly with Harmankaya et al. (2025). The MSDV is calculated for all conditions: on-road, test-track and simulator. The MSDV in the Simulator condition is almost 9-10 times lower than the On-road and Test-track conditions. This difference in the MSDV is caused by the high differences identified in the low-frequency accelerations, which are the most provocative to carsickness. Despite these differences in MSDV, the analysis of the subjective motion sickness data is important to explore how the signal has perceived by the participants.

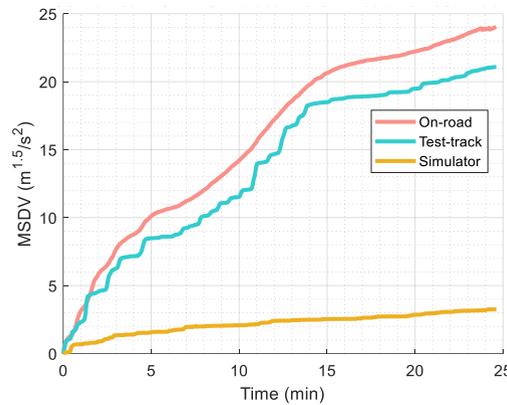

**Figure 3:** The calculation of the MSDV [$m^{1.5}/s^2$] for on-road, test-track and simulator data. The calculation was conducted based on ISO-2631 and as Harmankaya et al. (2025)

**Subjective data analysis:** In this section, the MISC values are plotted and analysed per condition (Figure 4). The mean MISC values across all participants per condition are examined are plotted over the time of the experiment. By the end of the experiment, the mean maximum MISC values are 2.74 for the on-road condition and 2.34 for the test-track condition, compared to only 1.07 for the simulator. This is a reduction difference of ~61 % (towards on-road) and ~54 % (towards test-track). According to the mixed-design ANOVA (LME model), there was a significant main effect of Time ($F(1, 3114) = 714.84$, $p < .001$), indicating a general increase in motion sickness symptoms over the session duration. Crucially, a significant Environment x Time interaction was observed ($F(2, 3114) = 47.71$, $p < .001$). Based on Figure 3, there are no differences between the three groups during the initial stages of the sessions (Minutes 0–8), but a clear divergence in symptom intensity emerges as time progresses. However, significant differences between the Simulator vs. Test-track and On-road first appeared at Minute 13[th] ($p = 0.012$) and 14[th] respectively ($p = 0.014$). Then, until minute 19[th], the magnitude of the difference increases, with the data illustrating significant differences.

However, between 20th–23rd minute, the MISC scores are again not significantly different. Despite this, the final MISC concludes to be significantly different between the Simulator and the other two.

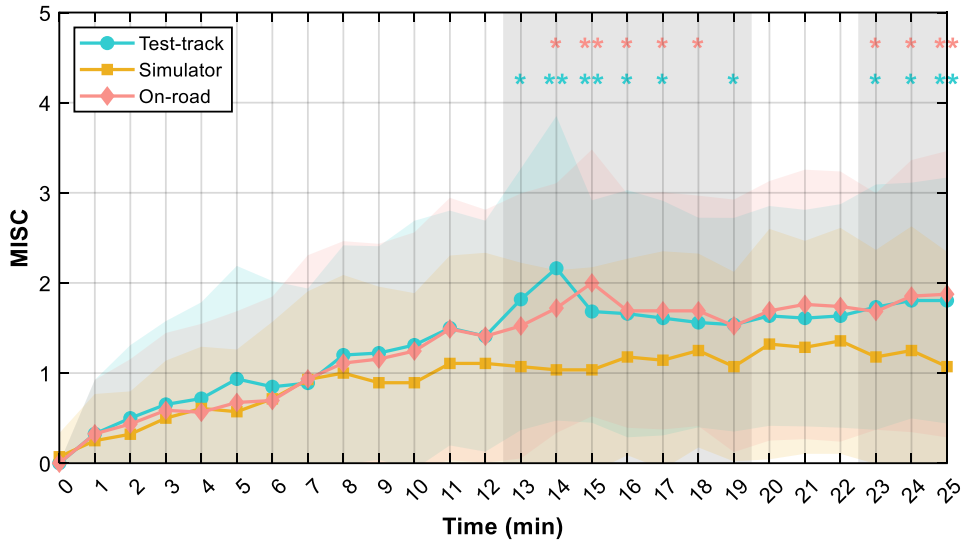

**Figure 4:** Mean and standard deviation of MISC. Test significance based on post-hoc analysis using series of independent Mann-Whitney U tests (rank-sum tests) at each minute is presented by ∗ p ≤ .05, ∗∗ p ≤ .01, and ∗∗∗ p ≤ .001

## DISCUSSION

In this section, we will discuss the results of the objective and subjective data analysis by exploring the replication of the on-road sickness exposure by the driving simulator.

<u>Objective motion sickness assessment:</u> The limited workspace envelope of the motion platform leads to reduced values of the MSDV, primarily because the platform cannot reproduce low-frequency motion components with sufficient amplitude. Low-frequency cues typically require large actuator strokes, which in this case significantly exceed the available travel of 670 mm. As a result, the motion cueing algorithm tries to wash out these components, leading to an underrepresentation of sustained accelerations in long-period motions.

This limitation is well known in the field of driving simulators and is commonly addressed through tilt coordination, where low-frequency translational accelerations are partially substituted by platform tilt. While this technique can effectively extend the motion perception of sustained accelerations, it introduces additional sensory cues that can be perceived by participants, inducing sensory conflicts. For this reason, in our study, the amount of tilt coordination was constrained to remain below human perception thresholds (Colombet, 2017), ensuring that participants did not detect artificial tilt cues. Nevertheless, restricting the tilt magnitude further reduces the achievable motion fidelity in the low-frequency domain, which contributes to conservative MSDV and limit the ability to fully reproduce real-world driving dynamics. A more advanced solution would be the use of a 9 DoF driving simulator architecture, in which the hexapod is mounted on a larger motion system, such as a tripod or a cable-driven platform (Cheli, 2022). This configuration enables substantially larger translational motion, allowing improved reproduction of low-frequency motion components.

**Table 1.** P-values for the test significance based on post-hoc analysis using series of independent Mann-Whitney U tests (rank-sum tests between Test-track vs. Simulator, and On-road vs. Simulator) at each minute. The shading cells correspond to the ones with significant difference ($p < 0.05$).

| Minute | Test-track vs Simulator | On-road vs. Simulator |
|---|---|---|
| 1 | 0.628 | 0.628 |
| 2 | 0.601 | 0.780 |
| 3 | 0.730 | 0.985 |
| 4 | 0.892 | 0.303 |
| 5 | 0.414 | 0.895 |
| 6 | 0.828 | 0.468 |
| 7 | 0.891 | 0.535 |
| 8 | 0.435 | 0.865 |
| 9 | 0.160 | 0.353 |
| 10 | 0.187 | 0.267 |
| 11 | 0.101 | 0.236 |
| 12 | 0.181 | 0.248 |
| 13 | 0.012 | 0.131 |
| 14 | 0.002 | 0.014 |
| 15 | 0.004 | 0.002 |
| 16 | 0.039 | 0.030 |
| 17 | 0.024 | 0.020 |
| 18 | 0.065 | 0.048 |
| 19 | 0.022 | 0.078 |
| 20 | 0.061 | 0.109 |
| 21 | 0.063 | 0.077 |
| 22 | 0.095 | 0.120 |
| 23 | 0.021 | 0.025 |
| 24 | 0.014 | 0.016 |
| 25 | 0.005 | 0.004 |

<u>Subjective motion sickness assessment</u>: The high standard deviation of the average subjective carsickness is caused by the inclusion of susceptible and non-susceptible participants in this calculation. At the same time, the duration of the provocation had a significant effect on carsickness accumulation in the simulator condition. This result underlines the time-dependence development of motion sickness as presented across all testing environments (Pham Xuan, 2023, Pham Xuan et al, 2025). According to the results, the interaction between time and environment demonstrated that the progression of MISC over time was significantly influenced by the testing environment, providing robust statistical evidence that the simulator induced a significantly different sickness profile (i.e., lower levels) compared to the test-track and on-road conditions.

Interestingly, our analysis illustrated that the differences between the simulator and the other conditions became significant over time and as the amplitude increased. This might imply that the simulator data could be valid until a specific MISC level, which can be limited based on the simulator capabilities, i.e. workspace. Similar behavior was noticed by Talsma et al. (2023), where the simulator matched the on-road carsickness exposure for the first 5 mins. After this period, the divergence from the on-road exposure also occurred. Talsma et al. (2023) also illustrated significant differences between simulator and their "car" condition, which was in a closed test-track without any other users. Interestingly, Himmels et al. (2024) presented that MISC differs across simulators with more advanced simulators provoking less MISC on the same driving scenario, but this difference was also dependent on the scenario.

**CONCLUSION**

To sum up, this paper extended previous work by replicating on-road sickness exposure in a driving simulator and exploring the carsickness assessment across all testing environments (on-road, test-track and simulator). The assessment included objective (accelerations and MSDV) and subjective (MISC) data analysis. The results illustrated that the driving simulator was unable to replicate the on-road sickness exposure in low-frequencies resulting in 10 times lower MSDV compared to the on-road and test-track conditions. At the same time, the perceived carsickness based on MISC had smaller but significant differences across the testing environment. The significant differences rise from the moment the exposure reached higher amplitudes.

Further work is in progress to explore the differences in the carsickness assessment across testing environments in more depth and evaluating the differences between various psychological factors across testing environments.